# Using an 'energy pixels' approach to estimate the size a 2nd Generation DEL must be to train a high-resolution binding model

Authors: Boris P. Belotserkovskii, R. Edward Watts
Affiliations: Insitro Inc, 279 E.Grand Ave, South SanFrancisco, CA 94070

ABSTRACT
DNA-encoded libraries (DELs) are commonly used to find small-molecule binders to a target protein. But the hard part is no longer finding a hit. It is building a machine-learning model good enough to guide the medicinal chemistry that follows, one that can resolve the shift in affinity when a chemist changes a single atom.
Such a model needs two things from its training data: dense sampling of the small patch of chemical space around the hit, and an adequate number of genuinely tight binders within it. A diverse naïve DEL gives neither. It samples an enormous space, so any one neighborhood is covered far too thinly, and the tightest binders are rare. A model trained on that naïve DEL data must extrapolate to binders tighter than any it has seen, and that is where models can fail.
One way to get training data with both properties is a second-generation DEL built around the hit. We ask how complex that library must be. We treat binding as a sum of small, independent interactions, called "pixels" on the target and "antipixels" on the ligand. From the spread of affinities measured in the naïve library, with no structural information and nothing yet synthesized, we estimate the complexity such a library must have to supply that dense sampling and those tight binders.
The method does two useful things. It turns the number of tight binders a model needs into the number of compounds a library must contain. And it estimates the expected yield of a randomly designed second-generation DEL, so the gain from designing the library with machine learning or active learning can be measured against a baseline, without the expense of a control library as large as the experiment itself.

## 1. INTRODUCTION

Finding a molecule that binds a chosen protein used to be the hard part of early drug discovery. It is not anymore. DNA-encoded libraries have made it routine: one builds a large library of compounds, each tagged with a DNA barcode, washes the library over the target, keeps what sticks, and reads the barcodes of the binders by sequencing (6-8; reviewed in 9-14). Because the tags can be amplified and sequenced, libraries of billions of compounds can be searched at once. The same recipe of a large, diverse library undergoing selection is shared with phage display, ribosome display, mRNA display, and aptamer discovery methods (1-5). The bottleneck has moved downstream, to what one does with the hits.

What one wants is a model that can guide the medicinal chemistry that follows. In a medicinal-chemistry campaign a chemist changes a molecule by one or two atoms at a time and, before making anything, asks whether the change will help. Such a change can shift the affinity a great deal or hardly at all, and the model has to resolve which: a small shift the chemist can disregard, a large one the chemist needs to see coming. And it has to do so in a small patch of chemical space, the neighborhood of the current best compound and within the current lead series.

A model is only as good as the data beneath it, and fine resolution asks two things of that data. First, the data must sample the neighborhood densely. To resolve small differences between neighbors, the model must have seen many neighbors. Second, the data must include an adequate number of genuinely tight binders, because those are the compounds the campaign is trying to reach, and a model asked about them cannot be left to guess.

A naïve library gives neither. It spends a fixed number of compounds on an enormous chemical space, so any one neighborhood is covered far too thinly. And because it is not designed specifically for the target, its tight binders are rare. A model trained on such data, then asked to rank compounds tighter than any it has seen, is extrapolating, and extrapolation is where models can fail.

One way to get data with both properties is to build a second library focused on the neighborhood of a hit. This "second-generation" library keeps a chemical feature, or features, that the first-round hits share, and varies the rest. Because every member carries a feature that favors binding, the whole library is shifted toward tighter binders. Because it varies only the neighborhood of the hit, it samples that neighborhood densely. Such libraries are more work to build, but they are the kind of data a high-resolution model requires.

This raises a plain question: how complex must such a library be? How many compounds does it take to sample the neighborhood densely enough, and to hold enough tight binders, to train the model one has in mind? A library that is too small wastes the effort and still yields a model that cannot resolve what the campaign needs.

There is no easy way to get this fraction from first principles. One would like to treat all possible compounds as points in a "chemical space" and ask what fraction of that space binds the target, as was done for antibodies (15): a library is likely to contain a binder when its size, times the binding fraction, is about one. The logic carries over to small molecules unchanged. What also carries over is that there is no way to compute the binding fraction in advance, so instead of calculating it we measure it, reading it from the affinities of a library already in hand. Chemical space is hard to define, and the reactions used to make a library fix what structures are possible (16, 17); but these are features of any library built by selection, antibodies included, not obstacles peculiar to small molecules.

We take a different route. Rather than reason from first principles, we read the binding data already in hand, the affinities measured in the naïve library, and use them to estimate what a second-generation library made from the same chemistry will do. To make this precise we use an idea we call the **superlibrary**: the set of essentially all compounds a given chemistry could make. A naïve, generic (or non-targeted) library is a small random sample of it. From the affinities in the sample, we recover the affinity distribution of the whole, and from that we estimate the size a second-generation library must have to meet a model's needs.

The method does two things worth stating at the outset. It turns a required number of tight binders into a required number of compounds, so one can size a library to the model it is meant to train. And it gives the yield one should expect from a second-generation library whose varied parts are chosen at random, a number computed on paper with nothing synthesized. That random yield is a plain baseline against which to measure what machine learning or active learning adds when it designs the library instead. Measuring that gain would otherwise require building a random library as a control, at a cost as large as the experiment.

## 2. THE MODEL

### 2.1. General picture

The basic energy-pixel model is shown in Fig. 1.

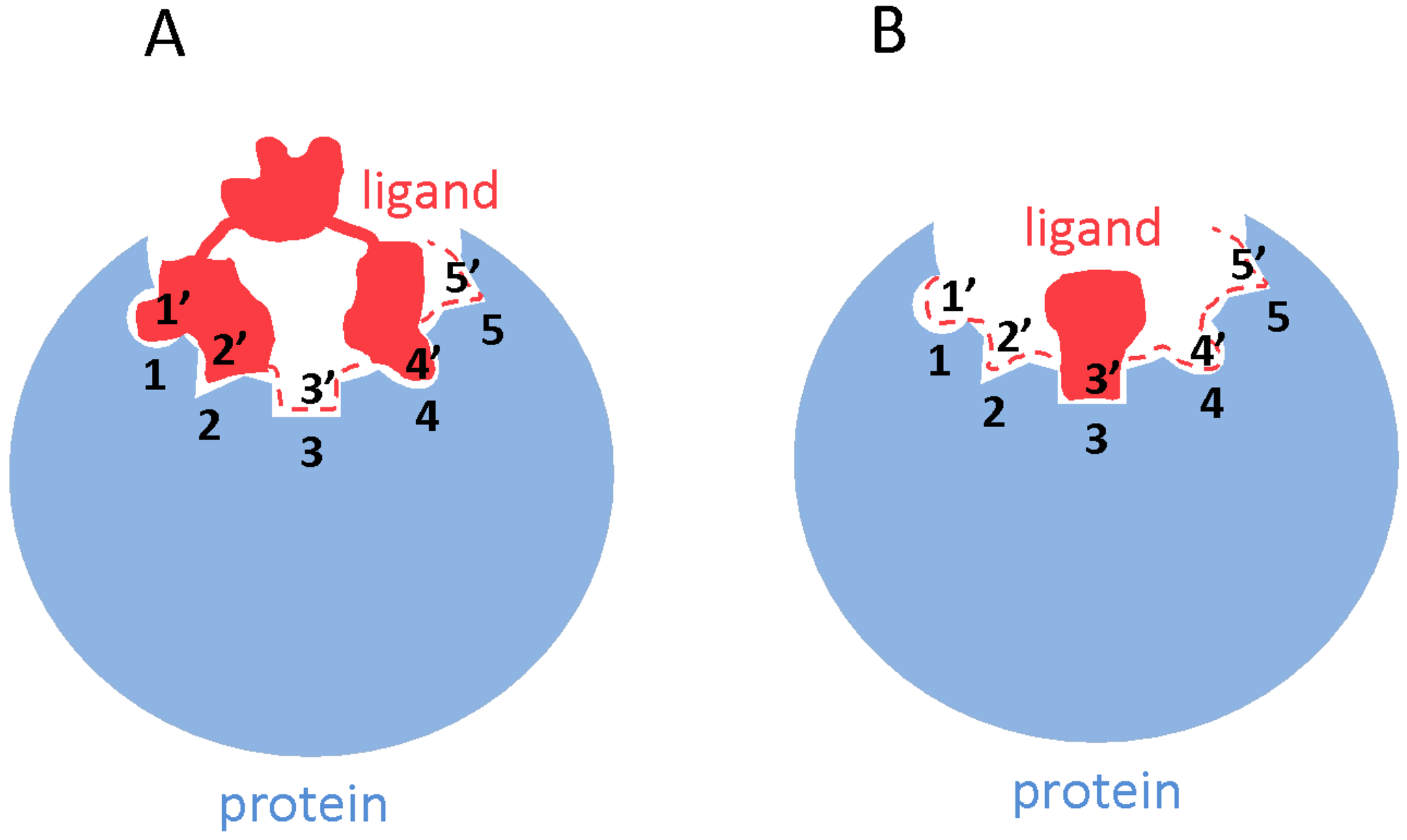

**Fig. 1. Ligand-target interactions in the energy-pixel model.**
Numbers 1 to 5 mark pixels; 1' to 5' mark antipixels. Dashed lines show "virtual" antipixels that are absent from the ligand and so contribute no interaction energy.

The target (blue) has a binding pocket dotted with binding spots, or “pixels,” numbered 1 to 5. The library compound, from here on the “ligand” (red), sits in the pocket in the position that makes their interaction as favorable as possible. In that position we picture the ligand as carrying “antipixels,” numbered 1’ to 5’, each facing the matching pixel on the target.

This does not mean a ligand must carry a real chemical group opposite every pixel. In Fig. 1A, pixels 3 and 5 have no ligand group near them; we simply say antipixels 3’ and 5’ contribute no energy. For the ligand in Fig. 1B only antipixel 3’ interacts at all; the rest are empty.

Each antipixel can take many chemical forms, each with its own interaction energy against its pixel. The superlibrary, in the ideal case, includes compounds carrying essentially every possible combination of antipixels, and the number of antipixels plays the part of the number of dimensions of chemical space. The total binding free energy (ΔG) between a ligand and the target is taken to be the sum of the free energies of all its pixel-antipixel pairs.

A pixel and an antipixel rarely match well, so for any given pixel only a small fraction of antipixels bind it strongly. It follows that within a superlibrary the tighter-binding ligands are the rarer ones (Fig. 2, top; there the fraction of binders, in gray, is drawn far larger than in any real library).

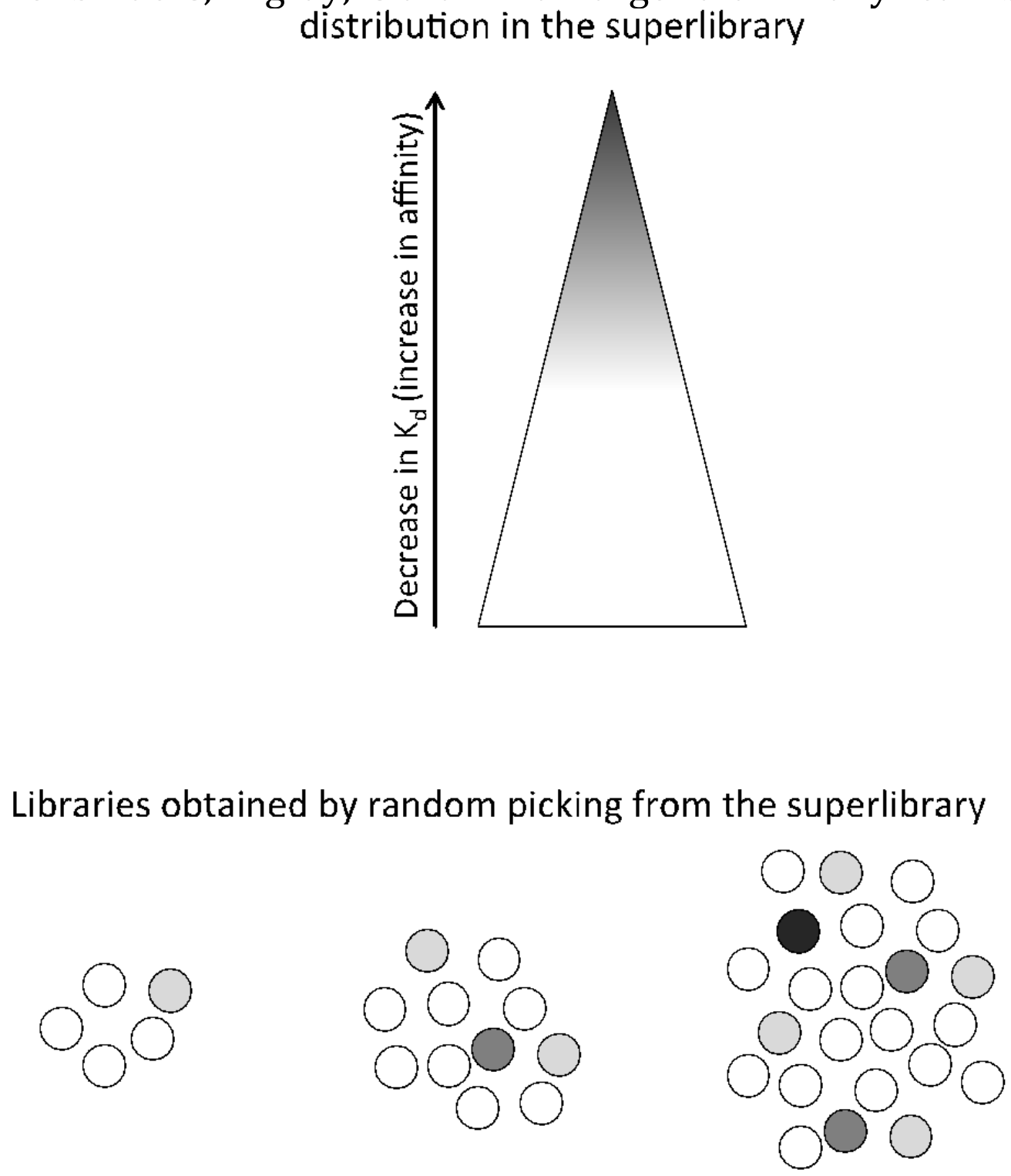

**Fig. 2. A combinatorial library in the energy-pixel model.**
Top: the binders within the superlibrary. Darker color means tighter binding; the narrowing width means rarer.

A real, finished DEL of complexity $C$ is treated as a random sample of $C$ compounds drawn from the superlibrary, whose complexity is far greater than $C$ (Fig. 2, bottom).

This applies to a whole naïve library and to any constrained part of it alike. A naïve library is a random sample of the whole superlibrary; a part of it that shares some constraint is a random sample of the part of the superlibrary that shares that constraint. The compounds in a naïve library that carry a phenyl group, for instance, are a random sample of all phenyl-bearing compounds in the superlibrary. The compounds sharing a constraint may make up a whole library or only a part of one; either way they are drawn from the same slice of the superlibrary and so share the same affinity distribution. This is what lets us predict the behavior of a second-generation library from a mere part of the first: the first-round hits that share a feature are a preview of the whole library one would build by fixing that feature.

And the rarer a property is in the superlibrary, high affinity say, the larger a library must be to contain a compound with it. We now make this quantitative.

Let $\nu$ be the chance that a ligand drawn from the superlibrary has some property, for example a dissociation constant in a chosen range. Then the chance that a library of complexity $C$ contains exactly $m$ such ligands follows the binomial distribution:

$$W(m, C) = \frac{C!}{(C-m)!\,m!}\nu^m(1-\nu)^{C-m} \approx \frac{(\nu C)^m}{m!}e^{-\nu C} \qquad (1)$$

The right-hand side is the Poisson approximation, which holds as $C$ grows large and $\nu$ grows small while their product $\nu C$ (the mean value of $m$) stays finite.

The chances that no such ligand is present, and that at least one is, are

$$W(0, C) = (1-\nu)^C \approx e^{-\nu C} \qquad (2)$$

and

$$1 - W(0, C) = 1 - (1-\nu)^C \approx 1 - e^{-\nu C} \qquad (3)$$

(One can also read $\nu$ as the chance that a ligand falls within a certain region of chemical, or shape, space. Perelson and Oster (15) treated all possible antibodies as filling a large but finite region of $N$-dimensional shape space, within which the binders occupy a small sub-region of relative width $\hat{\varepsilon}$ along each dimension, so that $\nu = \hat{\varepsilon}^N$.)

From Eqs. 2 and 3, the condition for a good chance, about 63 percent or better, of finding such a ligand in the library is

$$\nu C \gtrsim 1 \qquad (4)$$

We use this to find the complexity a library needs to contain a ligand tighter than some chosen affinity, that is, one with a dissociation constant $K_d$ below a chosen value $K_{d(min)}$. Suppose the fraction of superlibrary ligands with $K_d$ below $K_{d(min)}$ is $\Omega\left(K_{d(min)}\right)$. Then a library of complexity $C$ is expected to contain $\Omega\left(K_{d(min)}\right)C$ of them. If that product is well below one, such ligands are unlikely to be present; if it is well above one, they are very likely present, and there will be many. The middle case, $\Omega\left(K_{d(min)}\right)C \approx 1$, marks the presence of about one ligand tighter than $K_{d(min)}$.

Our aim is to obtain $\Omega\left(K_{d(min)}\right)$ from the affinity distribution of the naïve library, or of a constrained part of it, and from it estimate the complexity a second-generation library needs.

### 2.2. General mathematical treatment

In the model, pixel-antipixel pair $i$ has interaction energy $\Delta G_i$ with probability $p_i(\Delta G_i)$. In general, these probability functions differ from pair to pair.

The chance $P(\Delta G, N)$ that the whole ligand has total binding energy $\Delta G$ is the convolution of the $p_i(\Delta G_i)$. To get it, one multiplies the $p_i(\Delta G_i)$ for all pixels and integrates the product over all pixel energies, holding the sum of the pixel energies equal to $\Delta G$:

$$P(\Delta G, N) = \int \prod_{i=1}^{i=N} p_i(\Delta G_i)\, d(\Delta G_i)\,;\ \ under\ condition \sum_{i=1}^{i=N} \Delta G_i = \Delta G \qquad (5)$$

This is the model in its most basic form. Section 2.3 takes up simplifications.

In most experiments $\Delta G$ cannot be measured directly. What can be measured is the equilibrium dissociation constant $K_d$ of the ligand-target complex. It is tied to the free energy of binding by

$$K_d = K_{d0}\, e^{\frac{\Delta G}{RT}} \qquad (6)$$

or

$$\Delta G = RT\, ln\, \frac{K_d}{K_{d0}} \qquad (7)$$

$K_{d0}$ is the dissociation constant at zero interaction energy. The dissociation constant is the concentration at which half the ligand is bound. With no interaction, "binding" is just two molecules sitting next to each other, and half-binding happens when the molecules are about their own size apart, a few nanometers, which corresponds to a concentration of roughly 1 M. So in most of our estimates we set $K_{d0} = 1$ M, though in finer work it could be fitted.

The chance that a ligand has $K_d$ between $K_{d1}$ and $K_{d2}$ is

$$\Phi(K_{d1}, K_{d2}) = \int_{RTln\frac{K_{d1}}{K_{d0}}}^{RTln\frac{K_{d2}}{K_{d0}}} P(\Delta G, N)d(\Delta G) \qquad (8)$$

Multiplying this by the library's complexity gives the predicted number of ligands in that $K_d$ range, a number one can measure, and so a way to test the model.

Of particular interest is the chance $\Omega\left(K_{d(min)}\right)$ that a ligand has $K_d$ below a chosen $K_{d(min)}$, that is, affinity above a chosen value:

$$\Omega\left(K_{d(min)}\right) = \int_{-\infty}^{RTln\frac{K_{d(min)}}{K_{d0}}} P(\Delta G, N)d(\Delta G) \qquad (9)$$

**2.3. The case of a Gaussian distribution over binding energies**

The model makes the ligand's binding energy a sum of independent pixel-antipixel energies, so the distribution of the ligand's binding energy is the convolution of the pixel distributions (Eq. 5). By the central limit theorem, the convolution of enough distributions is, under broad conditions, close to a Gaussian:

$$P(\Delta G) = \frac{1}{\sqrt{2\pi\sigma^2}} e^{-\frac{(\Delta G-\mu)^2}{2\sigma^2}} \qquad (10)$$

where $\mu$ and $\sigma^2$ are the mean and variance of the whole ligand's binding energy. (Since Eq. 10 does not show the pixel number $N$, we write $P(\Delta G)$ rather than $P(\Delta G, N)$; $N$ is still present, as a hidden parameter.)

The Gaussian has one great convenience: two parameters, a mean and a variance, easy to fit to data. It also has a limitation. A Gaussian runs to infinity in both directions, so it grants a non-zero chance to any binding energy at all, meaning that with a large enough library one could find a ligand of any affinity whatever. In reality, the binding energy is capped by the strength of the non-covalent forces between atoms, so there is a tightest possible affinity beyond which no amount of complexity helps. But in a real library the best ligands are likely to sit well short of that ceiling, and there the Gaussian should serve. We take it as a reasonable starting point.

The mean of the distribution is where the two kinds of library part company. For a naïve library, built with no target in mind, the mean sits near zero energy, $\mu \approx 0$ (red curve, Fig. 3). A second-generation library is built by fixing a feature that the naïve hits share, and that feature favors binding, so the whole distribution moves toward tighter binding, into negative energies, $\mu < 0$ (green curve, Fig. 3). This shift is the point of building the second-generation library: it is what makes tight binders common enough to sample densely.

Every experimental method has a floor of detection: below some stability the complex cannot be seen. So the distribution is observable only for binding energies below a certain value (blue dash-dot line, Fig. 3), set by the method. In Fig. 3 the observable and unobservable parts are drawn solid and dashed. The shaded areas under the curves are the cumulative chance of a ligand below a given energy threshold; for thresholds of any practical use these areas are far smaller than drawn, a millionth or a billionth of the total.

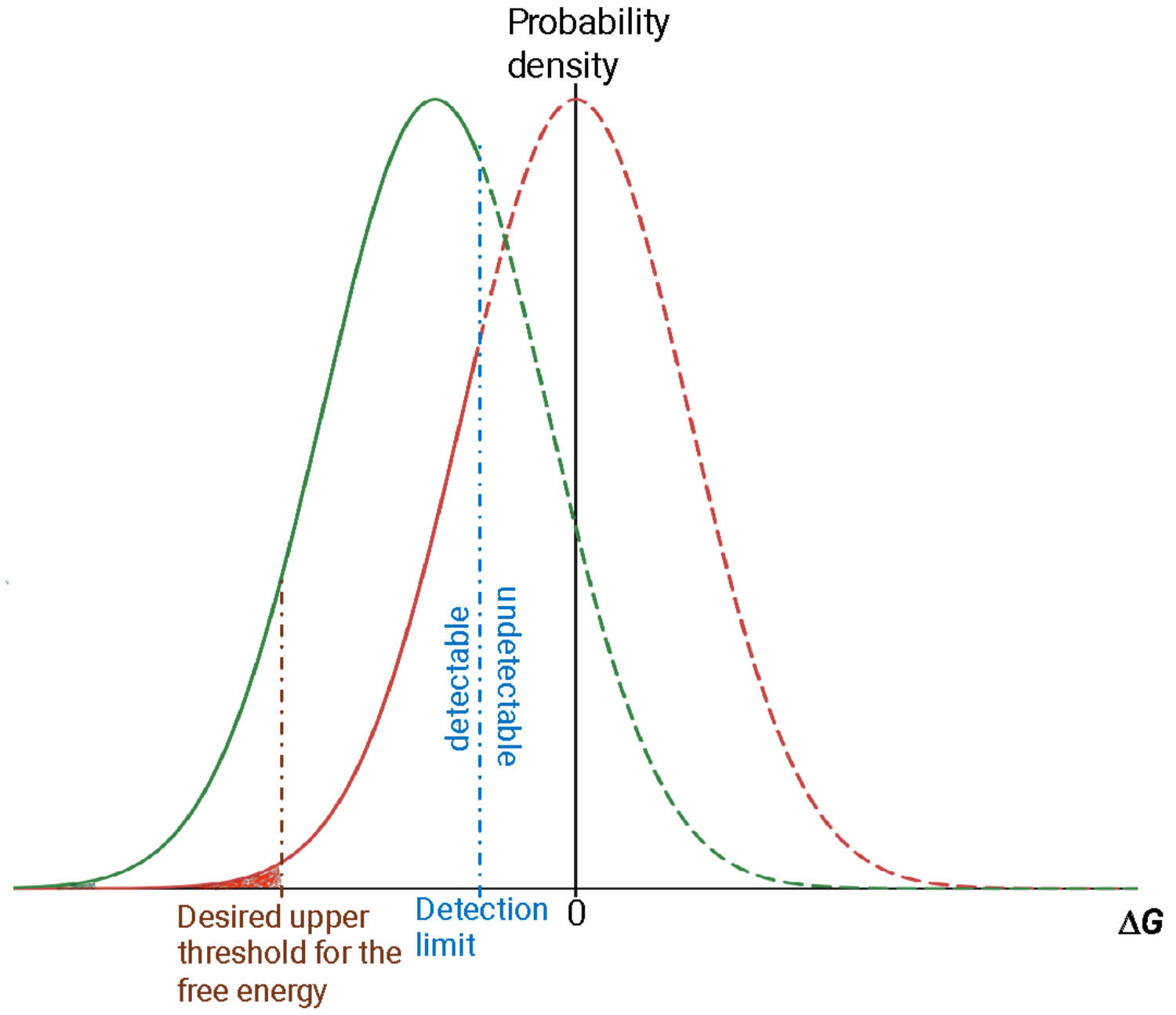

**Fig. 3. Distributions of binding energy in the Gaussian approximation.**
Red, a naïve library (mean near zero); green, a second-generation library shifted toward tight binders. Solid and dashed portions are the detectable and undetectable parts; the blue dash-dot line is the detection limit. See the text.

Putting Eq. 7 into Eq. 10 gives

$$P\left(ln\ \frac{K_d}{K_{d0}}\right) = \frac{1}{\sqrt{2\pi\sigma^{*2}}} e^{-\frac{\left(ln\ \frac{K_d}{K_{d0}} - \mu^*\right)^2}{2\sigma^{*2}}} \qquad (11)$$

where

$$\mu^* = \frac{\mu}{RT} \qquad (12)$$

and

$$\sigma^{*2} = \frac{\sigma^2}{(RT)^2} \qquad (13)$$

The chance that a ligand has a dissociation constant below $K_{d(min)}$ is

$$\Omega\left(K_{d(min)}\right) = \int_{-\infty}^{ln\ \frac{K_{d(min)}}{K_{d0}}} P\left(ln\ \frac{K_d}{K_{d0}}\right) d\left(ln\ \frac{K_d}{K_{d0}}\right) \qquad (14)$$

Putting Eq. 11 into Eq. 14 and setting

$$\frac{ln\ \frac{K_d}{K_{d0}} - \mu^*}{\sqrt{2\sigma^{*2}}} = \tau \qquad (15)$$

gives

$$\Omega\left(K_{d(min)}\right) = \frac{1}{\sqrt{\pi}} \int_{-\infty}^{\frac{ln\ \frac{K_{d(min)}}{K_{d0}} - \mu^*}{\sqrt{2\sigma^{*2}}}} e^{-\tau^2} d\tau = \frac{1}{2}\left(1 + \mathrm{erf}\left(\frac{ln\ \frac{K_{d(min)}}{K_{d0}} - \mu^*}{\sqrt{2\sigma^{*2}}}\right)\right) \qquad (16)$$

where

$$\mathrm{erf}(x) \equiv \frac{2}{\sqrt{\pi}} \int_0^x e^{-t^2} dt \qquad (17)$$

The complexity $C$ a library needs to hold, with fair probability, at least one ligand with a dissociation constant below $K_{d(min)}$ follows from

$$C\,\Omega\big(K_{d(min)}\big) = 1 \tag{18}$$

or

$$C = \frac{1}{\Omega\big(K_{d(min)}\big)} \tag{19}$$

(see Eq. 4 and the text below it). For the Gaussian, Eq. 19 gives

$$C\big(K_{d(min)}\big) = \frac{2}{1 + \mathrm{erf}\left(\dfrac{ln\,\dfrac{K_{d(min)}}{K_{d0}} - \mu^*}{\sqrt{2\sigma^{*2}}}\right)} \tag{20}$$

In Fig. 4 we plot this in double-logarithmic form ($lg\ \ C$ against $pK_{d(min)} \equiv -\mathrm{lg}K_{d(min)}$, with $K_{d(min)}$ in mol/L). We set $K_{d0} = 1$ M (see below Eq. 7). The parameter $\mu^*$ is the shift of the distribution in units of RT (about 0.6 kcal/mol); its values are 0, −1, −2, −4, and −8 for the red, orange, green, blue, and violet curves. The value $\mu^* = 0$ is a naïve library; a more negative $\mu^*$ is a second-generation library shifted further toward tight binders. We set $\sigma^* = \frac{3.8}{\sqrt{2}}$, chosen so that the naïve curve ($\mu^* = 0$) requires a complexity of $10^9$ to reach $K_{d(min)} = 100$ nM, close to what some experiments show. (Some libraries, of course, match a given target better than others.)

The figure shows what the shift buys. A library shifted toward tight binders reaches a given affinity at far lower complexity than an unshifted one, or, read the other way, holds far more tight binders at a given complexity. That is why a focused second-generation library, and not a larger naïve one, is the practical way to get dense sampling of tight binders.

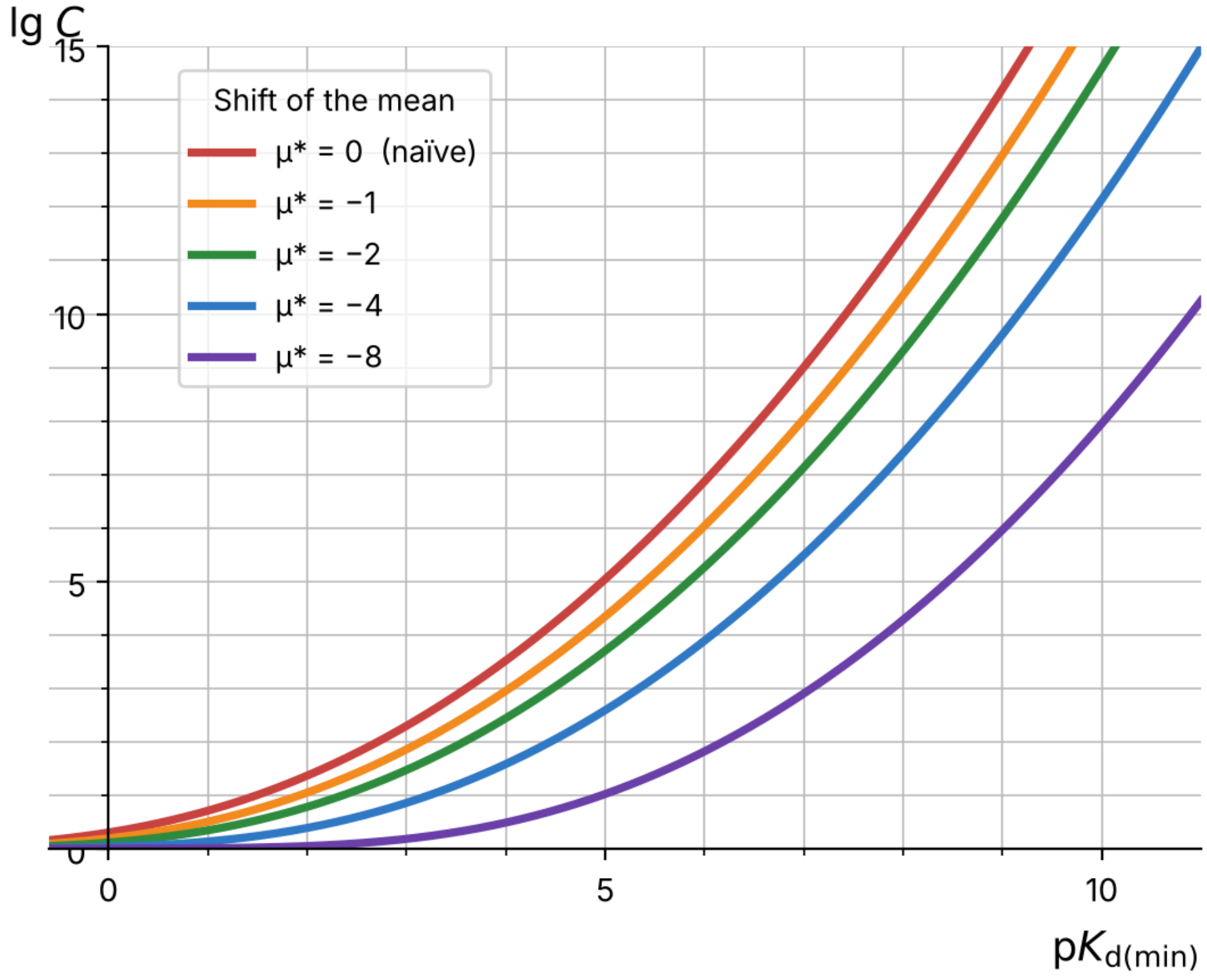

**Fig. 4. Complexity a library needs to contain a ligand with dissociation constant below $K_{d(min)}$, for several values of the shift $\mu^*$.**
$\mu^* = 0$ (red) is a naïve library; the more negative curves are second-generation libraries shifted toward tight binders. See the text.

### 2.4. Sizing a second-generation library to a model, and measuring what its design adds

So far, we have found the complexity needed for about one ligand below $K_{d(min)}$. A training set needs more than one. The expected number of ligands with $K_d$ below $K_{d(min)}$ in a library of complexity $C$ is $C\Omega\left(K_{d(min)}\right)$, as the counting above (Eqs. 3, 4, 8) already implies. To expect $n$ of them, set this equal to $n$:

$$C = \frac{n}{\Omega\left(K_{d(min)}\right)} = nC_1 \tag{21}$$

where $C_1 = 1/\Omega\left(K_{d(min)}\right)$ is the complexity that yields about one (Eq. 19). In plain terms, **needing $n$ tight binders multiplies the required complexity by $n$.** For the Gaussian, combining Eq. 21 with Eq. 20,

$$C = \frac{2n}{1 + \operatorname{erf}\left(\frac{ln\frac{K_{d(min)}}{K_{d0}} - \mu^*}{\sqrt{2\sigma^{*2}}}\right)} \tag{22}$$

This is the sizing tool. To use it for a second-generation library one needs $\Omega$ for that library, which means its affinity distribution, or in the Gaussian case its $\mu^*$ and $\sigma^*$. Here the argument of Section 2.1 pays off. The second-generation library will be a sample of the constrained slice of the superlibrary, the slice that shares the fixed feature. The naïve library's cluster of hits, the members that already carry that feature, is itself a random sample of the same slice. So its measured affinities estimate the distribution the second-generation library will draw from. One fits $\mu^*$ and $\sigma^*$ to those affinities and reads off the required complexity from Eq. 22. The number of members carrying a given feature is known from the library design, and so the number of those that did not bind within the detection limit is also easily counted. The fraction binding within the detection limit is therefore measured directly, and only the distance from the detection limit to the desired threshold has to be extrapolated, rather than the distance from the center of the distribution. Nothing need be synthesized to do this.

A word on the fixed feature, to make the shift concrete. Suppose the tighter hits in the naïve library tend to carry a phenyl group. Building every member of the second-generation library with a phenyl fixes that feature, and because the feature favors binding, the whole distribution shifts toward tight binders. If a methylated phenyl favored binding still more, and the hits pointed to that, then fixing the methylated phenyl would shift the distribution further. The measured affinities of the cluster of hits carry this information already, which is why they, and not a guess, set $\mu^*$ and $\sigma^*$.

The same expected count, $C\Omega\left(K_{d(min)}\right)$, is the yield one should expect from a second-generation library whose varied parts are chosen **at random**, because Eq. 21 assumes exactly that, a random sample of the constrained slice. This gives a plain benchmark. A library designed by machine learning or active learning is meant to do better than random: to choose the varied building blocks so as to enrich for tight binders beyond a random fill. Its worth can be read against the random baseline in two equal ways.

At a fixed library size, compare the number of tight binders the designed library actually yields with the expected random number $C\Omega\left(K_{d(min)}\right)$. The ratio of the two is what the design bought. A ratio near one means the design did no better than chance.

At a fixed target affinity, compare the size the designed library needed to supply $n$ tight binders with the random size from Eq. 22. A smaller size is the saving the design delivered. To measure what a design adds, one would normally build both the designed library and a random one, screen both, and compare. The random library is a control, and it costs as much as the experiment. **Here the control is replaced by a calculation.** One builds and screens only the designed library and measures it against a random baseline worked out in advance from the naïve library's own data.

This baseline is a cheap and honest yardstick, and the sizing of Eq. 22 is a way to build a library to the resolution a model needs rather than to a guess. It rests on assumptions, though, and the next section is plain about them.

## 2.5. What the estimate assumes

The estimate rests on a few assumptions, and it is worth being plain about them. One is mild: that the varied parts of the second-generation library are a fair random sample of the constrained slice of the superlibrary, which is what lets the naïve library's cluster of hits stand in for the library one has not yet built. When a design departs from random on purpose, as machine learning and active learning do, the random figure is simply the baseline it aims to beat.

The key assumption is additivity. The pixel model takes binding energy to be a sum of independent pixel contributions, and that independence is the same assumption that sits, in its simplest form, under many models built on molecular fingerprints and graph neural networks. Such models can in principle learn non-additive relationships between groups, but in typical practice, when they work it is largely because binding is close to additive, and when they fail it is often because it is not. In our estimate the assumption enters at one point. Fixing a feature restricts the library to the sub-library that carries it. That sub-library has a shifted mean and a narrower width than the whole library, but its distribution is the same in the naïve DEL, the superlibrary, and the second-generation DEL. This holds when a feature contributes the same amount whatever else is present, so that fixing it adds a constant to a sum over the still-random remaining pixels, and the distribution slides toward tighter binding and narrows. It fails when features cooperate, when the contribution of one depends on what the others are doing; then fixing a feature changes the shape of the distribution and not just its mean, and the cluster of hits no longer predicts the second-generation library by a simple shift. Some of this can be absorbed by treating a cooperating combination as a single unit, a lumped pixel, which restores additivity when the combination acts all-or-nothing and on its own; it does not help when the coupling is graded or spread across many features, and a combination that must be present as a whole is rarer than its parts, so the complexity the method returns rises accordingly. (Free and Wilson, in the very paper that introduced the additive treatment of substituent effects (18), found that a more elaborate model added nothing to their series. Additivity is often enough, and the data tells you when it is not.)

A separate hazard sits at the same spot, and it applies even when additivity holds. We see only the detectable tail of the distribution: the members that carry the feature and also bind tightly enough to be measured. Fitting the mean and width of the whole distribution from that tail is an extrapolation, and it grows less trustworthy the further the desired affinity lies beyond what was actually measured. Deeper and more accurate affinity measurement is the direct remedy.

Neither hazard is fatal. When additivity holds, the estimate is more accurate; when it fails, less so. The additive pixel model is a first-order tool for estimating how much data to gather, and a richer model is what exploits the places the first-order picture breaks. (Where additivity fails only mildly, the per-feature contributions can in principle be recovered from the same data and the estimate corrected; we do not pursue that here.) The point to hold onto is that the estimate is not a guess. Before committing the time and money a second-generation library costs, one has a clear figure for how large it needs to be in order to meet the needs of a model driving a medicinal chemistry campaign.

## 3. DISCUSSION

This model estimates the complexity a combinatorial library needs to supply what a high-resolution binding model requires: dense sampling of the neighborhood of a hit, and an adequate number of tight binders within it.

It treats the interaction between a ligand and its target as a sum of small, independent interactions, "pixels" on the target and "antipixels" on the ligand (Fig. 1), and takes the total binding energy as the sum over all pixel-antipixel pairs. Adding up the contributions of separate chemical groups this way is common in structure-activity modeling (18). But in our model an antipixel need not be a particular chemical group in the ligand. In general, it is an apparent "unit of interaction" whose parameters are found from experiment. And as Fig. 1 shows, the antipixel description of a ligand depends on the shape of the target's binding site: the same ligand may be described differently against different targets. The model is therefore written in the frame of a particular target, and the same library will in general be described differently for different targets.

The pixel approach earns its keep in two ways, both aimed at the predictive model a campaign is trying to train. First, it sizes the library. Such a model must call very tight binders correctly, and it can only do so if it has been trained on an adequate number of very tight binders. Those binders are exactly what a naïve library is poorest in, and what a model reaching past its training data can't achieve by wishful extrapolation. It is worth naming what shifts between the training data and the compounds the model must judge. The inputs barely move because the second-generation library is built from the same chemistry as the first, the two hold much the same kinds of molecule, so covariate drift, a shift in the inputs, is small, and we treat it as negligible. What moves, and moves a great deal, is the binding affinity itself, the quantity being predicted. That is prior shift, or label drift, and here it can be severe, which is precisely why a model must have seen real tight binders and should not be trusted to invent them. The relation $C = n/\Omega$ (Eq. 21) turns "an adequate number of very tight binders" into a required library size, so the pixel approach tells us how complex a second-generation library must be to train a reliable, medicinal-chemistry-worthy model, not merely to yield a single hit.

Second, it supplies a benchmark. Because the expected result of a randomly designed second-generation library can be computed from the naïve library's affinity distribution (Section 2.4), the gain from machine learning or active learning can be measured against a calculated baseline, without building a random control library.

The method is not tied to DNA-encoded libraries, nor to binding. A DEL is only a convenient way to make a large library and read a property from it; the estimate itself needs just two things, a library that can be treated as a random sample of some larger space of possible compounds, and a property distributed over that space whose distribution can be estimated, and shifted predictably, from data already in hand. Read this way, the same estimate that sizes a library for a binding model also sizes one for a model of solubility, permeability, or any other structure-dependent quantity a campaign must predict.

Because the model works with binding free energies, which are tied directly to equilibrium dissociation constants, fixing its parameters calls for high-throughput measurement of the dissociation constants of library members, and across the tight-binding range that the model cares about. This is not a small requirement: the same tight binders the downstream model needs are the ones that let this model be parameterized in the first place. Dissociation constants are not usually measured during a DEL selection, but selection procedures we and others have described (19-21) can be adapted to yield dissociation constants at the scales this type of model needs in order to be tested and used.

Acknowledgments: Gratitude is due to Christoph Klein and Nate Stanley for their edits to the manuscript, and to Spurti Akki, Tomek Danel, Gabriel Dreiman, Victor Galievsky, Kent Gorday, Divya Kanichar, Vincent Lee, Patrick J. McEnaney, Kirill Novikov, Virja Pandya, Nathaniel Stanley, Sophia Tabchouri, Joshua Turnbull, and Jason Wang for their experimental and computational efforts.

Conflict of interest statement: BB and REW are employed by or paid consultants of Insitro Inc.